\documentclass{article}
\usepackage{ijcai23}

\usepackage{times}
\usepackage{soul}
\usepackage{url}
\usepackage[hidelinks]{hyperref}
\usepackage[utf8]{inputenc}
\usepackage[small]{caption}
\usepackage{graphicx}
\usepackage{amsmath}
\usepackage{amsthm}
\usepackage{booktabs}
\usepackage{algorithm}
\usepackage{algorithmic}
\usepackage[switch]{lineno}

\title{Inductive Models for Artificial Intelligence Systems are Insufficient without Good Explanations}

\author{
    Udesh Habaraduwa
    \affiliations
    Tilburg University School of Humanities and Digital Sciences : Department of Cognitive Science and Artificial Intelligence
    \emails
    u.habaraduwakandambige@tilburguniversity.edu
}

\begin{document}

\maketitle

\begin{abstract}
This paper discusses the limitations of machine learning (ML), particularly deep artificial neural networks (ANNs), which are effective at approximating complex functions but often lack transparency and explanatory power. It highlights the `problem of induction'—the philosophical issue that past observations may not necessarily predict future events, a challenge that ML models face when encountering new, unseen data. The paper argues for the importance of not just making predictions but also providing good explanations, a feature that current models often fail to deliver. It suggests that for AI to progress, we must seek models that offer insights and explanations, not just predictions.
\end{abstract}

\section{Introduction}

The recent and unprecedented successes of extremely high-dimensional (i.e., over-parameterized) deep neural network methods (and ML in general) in language processing \cite{vaswani2017attention}, image generation \cite{ho2020denoising}, reasoning \cite{bubeck2023sparks}, and numerous other tasks have heralded what seems to be another `spring' in artificial intelligence (AI) research. Fueled by colossal data sets and an increasingly powerful, distributed, and reliable computing infrastructure, these models have been able to \textit{approximate solutions} to many problems considered too difficult to solve. Nevertheless, with great power comes great responsibility. These advances have been shadowed by escalating concerns over the transparency, explainability, and interpretability of these `black box' models \cite{guidotti2018survey,rudin2019stop}. Conversely, some argue that the learned parameters of these models should be regarded as scientific theories in their own right \cite{piantadosi2023modern}. How, then, can we reconcile the complexity of these models with the need for comprehensibility? Is it possible to train models of arbitrary complexity that are still capable of elucidating their own operations ?

This paper posits that the predominant challenges in \textit{understanding} how ML models function, and why they often go horribly wrong \cite{selbst2019fairness,zhang2021understanding,o2017weapons,szegedy2013intriguing}, is deeper than AI and that it is rooted in the false promise of (machine) induction \cite{hume2016enquiry}, and models developed through induction (i.e., ML, neural or not), as a method of understanding a given phenomenon. Instead of providing an \textit{explanation} of a phenomenon, models trained this way present us with yet another phenomenon that needs an explanation \cite{wiegreffe2019attention,jain2019attention}. Thus, despite the recent surge in the field of `explainable AI' \cite{doshi2017towards}, which attempts to provide some insight in to the generalizations made by trained models, it may be the case that the underlying \textit{problem of induction} and a lack of \textit{good explanations} will remain so long as we use machine induction as the primary path in AI.

\section{The Problem of Induction}

How do we generate (new) knowledge? Generally, the inductive method of knowledge generation, championed by Francis Bacon \cite{bacon1878novum} in the seventeenth century, involves the collection of empirical data through observation and experimentation. Baconian induction suggests that from these observations, general laws and theories can be derived.
The problem of induction, as articulated by David Hume in the eighteenth century, challenges the legitimacy of inferential reasoning from the observed to the unobserved \cite{hume2016enquiry}. Hume’s skepticism about induction posits that, for example, just because the sun has risen every day in the past, there is no necessary reason to conclude it will continue to do so. Indeed, consider that no human has ever observed the death of our own sun yet we are able to predict that it will one day cease to rise billions of years in the future \cite{hansen2012stellar,sackmann1993our}.Previous observation alone would not be enough to predict this eventual outcome .

In the context of ML, this philosophical quandary manifests in the assumption that models trained on historical data will generalize to unseen data—effectively predicting the future based on the past. That is, a ML model (e.g., a neural network or linear regression) adjusts it's internal parameters progressively over repeated exposure to examples until it's predictive performance reaches an acceptable standard (e.g., classification accuracy). The problem begins with the sample of examples collected. Indeed, it is impossible to collect all possible samples (e.g., the space of all possible sentences is infinite) thus a \textit{representative} sample is sought. Here, representative is taken to mean a sample such that future samples are unlikely to deviate significantly from the past. Therefore, all inductive models reserve the possibility of failure should they be presented with an example that, though valid, was unobserved in the training sample. The ML literature is replete with examples where model performance plummets when out-of-distribution test cases are introduced (e.g., slight variations are made to medical images \cite{beede2020human} or a game environment is minimally altered \cite{kansky2017schema}). 

In a broader sense, the types of models that are built through induction and the types of models (i.e., explanations) that allow us , for example, to predict the death of a star or the bending of light around massive objects \cite{dyson1920ix,einstein1922grundlage} without ever having experienced such phenomenon, both engage in a form of prediction \cite{elton2021applying}. The latter however is not only more robust to as yet un-sampled examples but some how \textit{predicts that these examples will exist} \cite{deutsch2011beginning}(e.g., general relativity predicted black holes before any were ever observed \cite{akiyama2019first}). While inductive methods are able to `generalize' within distribution, our best \textit{explanatory models} are able to extrapolate across distributions\cite{elton2021applying}. What distinguishes the former from the latter?

\section{A good explanation}

Building on the work of David Hume, Karl Popper further refined our best known process of knowledge generation, the scientific method, as the iterative process of generating ever better theories. The hallmark of a scientific theory is not its ability to be confirmed by observational data (as inductivism would suggest), but its capacity to be falsified and withstand attempts at refutation \cite{popper2005logic}. A defining difference between this framing and that of Bacon's is the role of conjecture. For Bacon, knowledge generation begins with the `unbiased' collection of data and then proceeding to derive knowledge about the general from what is observed in the specific \cite{bacon1878novum}. Conversely for Popper, bold conjectures (i.e., hypotheses) are made about the world, tests are formulated which, if failed, would eliminate a given conjecture from consideration as a valid \cite{popper2014conjectures}. Through an evolutionary process of generation and elimination, the body of knowledge is continually refined as we move away from bad explanations to better ones \cite{popper1979objective}.

Is falsifiability alone a sufficient criterion for a conjecture to be a candidate for a scientific theory ? To address this problem, the physicist and philosopher David Deutsch proposed the criterion of being `hard to vary' \cite{deutsch2011beginning}. Consider a conjecture that is sufficiently vague, like that of an astrologist, Marxism, and some psychological theories \cite{popper2014conjectures}. These retain a facade of `robustness' by the fact that they can morph into whatever shape is necessary to better explain observations placed in front of them. Additionally,they may also contain components that can be swapped out for any other. For example, if we attempt to explain the changing of seasons as the acts of the goddess Demeter and a marriage contract of her daughter to Hades \cite{deutsch2011beginning}, there is no reason we can't swap in any other deity from the pantheon or a different agreement, while maintaining the theory's explanatory power.

A scientific conjecture is not only falsifiable but also (ideally) a good explanation: that is, it is a risky prediction \cite{deutsch2011beginning,popper2014conjectures}. The degree to which contradictory evidence is destructive to a conjecture is a function of how precisely it is defined. A precisely defined conjecture cannot be easily altered to accommodate new evidence, making it brittle in the face of evidence that is contradictory (i.e., a risky prediction). It is this property of scientific explanations that gives them their `reach' : their ability to extrapolate across distributions which inductive methods (e.g., deep learning, linear regression, etc.) cannot manage \cite{deutsch2011beginning,elton2021applying,nielson2021induction}. 

\section{Good enough predictions over good explanations}

The world is a complex place. Many of the phenomenon that humans experience (e.g., qualia \cite{kanai2012qualia}) and actions we take as we navigate our environment remain largely a mystery to us. For example consider the act of driving a car. This is a task that is accomplished by billions of people around the world. Though on its face, this is a simple learning task that we entrust many teenagers with, when trying to imbue a machine with this ability, we find a web of difficult, interconnected decisions, from the ethical (e.g., the trolley problem \cite{thomson1984trolley}) to the physical (e.g., turning angle) , that a machine must successfully navigate \cite{maurer2016autonomous}. Curiously, one does not encounter an evaluation of moral quandaries like the trolley problem in a driving exam; rather, the driving exam focuses on practical aptitudes and a degree of faith on part of the examiner (and the public) that the driver will `do the right thing'. 

Thus arises the question: How do humans, as flawed as we are, manage to `do the right thing' on the road? What is the \textit{process} that we employ to accomplish this task? Historically, this was a pivotal question because if we were to get a machine to perform any task, we would need to program it with a set of instructions to do so. At the inception of the field, it was presumed that many aspects of human intelligence could be carefully described with enough precision as to be implemented in a computer. With this intuition in hand, researchers set out to elucidate and then implement reasoning and decision making algorithms in machines (e.g., classical search strategies \cite[Chapter 4, 5, and 6]{russell2016artificial}, inference in first-order logic \cite[Chapter 9]{russell2016artificial} , etc.) in an attempt to emulate some of the behaviors that humans effortlessly produce. However, as the pioneering group gathered at Dartmouth on the summer of 1956 soon found out \cite{mccarthy2006proposal}, this would be a much more difficult task than they had imagined. Attempts at emulating human performance in a variety of domains (e.g., language comprehension \cite{winograd1972understanding}, common sense reasoning \cite{lenat1990cyc}, and motor control \cite{moravec1983stanford} ) in this era, relying heavily on symbolic processing and rule-based systems as they did, made little headway. This era of AI research (1950-1960) has come to be known \textit{and taught} as the era of `good old fashioned artificial intelligence (GOFAI)' \cite{newell2007computer,haugeland1989artificial}. Faced with the daunting task of developing good explanations, the field turned to the next best thing : a good enough approximation.

The earliest method of formalizing induction, the humble least-squares method, was developed independently in the nineteenth century by Carl Friedrich Gauss \cite{gauss1877theoria} and Adrien-Marie Legendre \cite{legendre1806nouvelles}. The method provided a powerful tool for modeling and, perhaps most importantly, predicting the future value of a dependent variable. In its earliest use by Gauss, the least-squares regression method was used to predict the position of the asteroid Ceres at some point in the future. In 1801, the Italian astronomer Joseph Piazzi was tracking Ceres and managed to make 22 observations until it was lost to sight against the light of our sun. To find it once it reappeared in the sky, astronomers would have to calculate Ceres' orbit or find another way to predict its future position \cite{buhler2012gauss,gauss1877theoria}. Armed with an equation of six elements (derived from Kepler's laws of planetary motion) Gauss used Piazzi's observations and iterative changes to the model parameters to predict where Ceres would be along its orbit \cite{gauss1877theoria}. In the present context, there is a distinction that should be drawn here between Gauss's \textit{prediction} and Kepler's \textit{explanation} of planetary motion. The former, while it provides a mathematical tool for prediction, provides no theoretical explanation of the movements of celestial bodies (i.e., a good explanation) like the latter which gives us insight into the fundamental principles governing planetary motion. The term `regression' was later coined by Francis Galton \cite{galton1886regression} when he calculated the `line of best fit' between the height of parents and their offspring, inducing a pattern for the general from a specific sample. Galton's statistical work on heredity and the later formalizations developed by his protege Carl Pearson \cite{pearson1894contributions,pearson1901liii} laid the ground work for what is now much of modern statistical theory. Importantly, this would herald the coming of the `model fitting' paradigm of science in general and later in the field of AI.

Linear regression by method of least-squares, defined as the process of finding the weights that, when applied to a set of variables, minimize the difference between the predicted and observed outcomes, is commonly taught as a foundational method under the umbrella of supervised learning algorithms. Faced with the limitations of linear models in predicting the outcomes of non-linear relationships (i.e., difficulty in approximating non-linear functions) a slue of ever more elaborate and effective models were developed (e.g., logistic regression \cite{cox1958regression}, non-linear regression \cite{cox1958regression}, decision-trees \cite{quinlan1986induction}, support vector machines \cite{cortes1995support}, etc.) that were able to induce excellent approximations to many non-linear solutions. It is on this path, owing to the inspiration drawn from the human brain \cite{mcculloch1943logical,rosenblatt1958perceptron} that we arrive at (deep) ANNs, the current state of the art in (supervised) machine induction. ANNs can be understood has extremely high-dimensional composite functions defined by the their internal parameters (i.e., the weights). Thanks to algorithmic breakthroughs (e.g., back-propagation \cite{rumelhart1986learning}) improved computing resources (i.e., machines optimized for matrix operations) and access to massive amounts of data (i.e., the internet), it became possible to do what Gauss did but approximate ever more elaborate functions. To make matters more interesting, the finding that an appropriately parameterized neural deep network is able to approximate any continuous function to an arbitrary degree of accuracy was a foundational result in the field \cite{cybenko1989approximation,hornik1991approximation}.

The beauty of ANNs (which are generally implemented as sequence of nested matrix operations) lies in the freedom they provide the designer to stack, combine, and remix its inner components (i.e., layers and neurons) in creative ways (e.g., convolutional ANNs \cite{lecun1989backpropagation}, Long short-term memory \cite{hochreiter1997long}, generative adversarial networks \cite{goodfellow2014generative}, attention \cite{bahdanau2014neural}, etc.) The core requirement is adherence to the principles of linear algebra and calculus, which ensure the propagation of errors backward through the network using gradient-based optimization methods. Thus, while Gauss was limited to approximating stellar orbits with the use of a function with only six free parameters, we are now able to approximate the likes of language comprehension\cite{vaswani2017attention} with a function of billions. The rest, they would have us believe \cite{bubeck2023sparks}, is history.

\section{We need better explanations}

The trouble with good predictions is that they are incredibly useful and seemingly easier to generate than good explanations. Despite the triumphs of ANNs and ML in approximating functions and predicting outcomes across various domains, the scientific community grows increasingly troubled, and rightly so, by lack of insight into the inner workings of these models. Though many claim that the problem is with `black box' ML models (e.g., deep ANNs), this unease stems from a fundamental limitation of these inductive approaches: they do not provide the explanatory depth that is characteristic of good scientific theories. Even those that argue for `explainable' ML methods \cite{rudin2019stop} nonetheless argue for \textit{explainable inductive models} that in no way guarantee the generation of a good explanation. In this sense, the hallucinations common to the current state of the art large language models \cite{zhang2023language} is not puzzling : it is a \textit{feature} of optimizing for the generation of plausible natural language, not a bug. The fact that it generates anything comprehensible at all is more a testament to the fact that human beings, for over 20,000 years, have systematically generated and stored information in a form amenable to computation \cite{WolframAlphaDocs}. 

Lacking any constraints, an ANN will `happily' find a way to optimize its objective function without necessarily `learning' representations that we would expect \cite{elton2020self,elton2021applying,ilyas2019adversarial,hasson2020direct}. Without any explanatory theory to bind it, we cannot assume that the internal mechanisms, regardless of what we analogize them as (e.g., learning, attention, short term memory, etc.), relates in anyway to what we actually want : a good explanation that can generalize to out of training distribution situations and/or represent a fundamental insight into phenomenon (be it language learning in children \cite{alishahi2010computational} or the orbit of an asteroid). The fact that a decision tree generates cut-offs for the variables it concerns doesn't \textit{explain why it chose to do so} :  it does not tell us why that feature is important or the causal relationship between the feature and the outcome. As with least-squares linear regression, we can see that one feature is given more weight than the other but the responsibility remains on us to come up with an explanation for it.  

\section{Conclusion}
Even though optimizing for predictive performance and model fitting has been incredibly useful, in terms of generating \textit{understanding} of the reality around us, there is a lot left to be desired (as highlighted by those raising the alarm about black-box models). All this is not to say that advances like that of the self-attention mechanism \cite{vaswani2017attention} are not significant breakthroughs in ML nor that ChatGPT is not a profoundly impactful product (indeed, it was used extensively to streamline the research process and to generate the abstract of the present paper). It is to say that they are nonetheless inductive models, like all supervised ML models, and will be subject to all the pitfalls inherent therein. To find the satisfaction that we hope to get from methods designed to investigate `black box' methods, we have to first recognize that we need good explanations and that inductively minimizing the loss in a prediction task might not get us there. We might also benefit from not teaching the early era of AI as `good-old-fashioned AI' and teach it as what it was : the search for good explanations of how human beings accomplish such a dizzying array of tasks and finding ways to implement them in machines.

\appendix

%% The file named.bst is a bibliography style file for BibTeX 0.99c
\bibliographystyle{named}
\bibliography{references}

\end{document}